\pdfoutput=1
\documentclass[11pt]{article}

\usepackage{ACL2023}
\usepackage{times}
\usepackage{latexsym}
\usepackage{threeparttable}
\usepackage{multirow}
\usepackage{amsmath}
\usepackage{graphicx}
\usepackage[T1]{fontenc}
\usepackage[utf8]{inputenc}
	
\usepackage{color, colortbl}
\definecolor{Gray}{gray}{0.9}
\usepackage{microtype}
\usepackage{pifont}% http://ctan.org/pkg/pifont
\newcommand{\cmark}{\ding{51}}%
\newcommand{\xmark}{\ding{55}}%
\usepackage{inconsolata}
\title{AFLoRA: Adaptive Freezing of Low Rank Adaptation in Parameter Efficient Fine-Tuning of Large Models}

\author{Zeyu Liu$^{\dagger,1}$ \
  Souvik Kundu$^{\dagger,2}$ \
  Anni Li$^{1}$ \ Junrui Wan$^{1}$ \ Lianghao Jiang$^{1}$ \ Peter A. Beerel$^{1}$ \\
  $^{1}$ University of Southern California, USA \ $^{2}$ Intel Labs, San Diego, USA \\
  \texttt{\{liuzeyu, annili, junruiwa, ljiang40, pabeerel\}@usc.edu} \ \texttt{souvikk.kundu@intel.com} \\
  $^{\dagger}$Equally contributing authors
    }

\begin{document}
\maketitle
\begin{abstract}
We present a novel Parameter-Efficient Fine-Tuning (PEFT) method, dubbed as \textit{Adaptive Freezing of Low Rank Adaptation} (AFLoRA). Specifically, for each pre-trained frozen weight tensor, we add a parallel path of trainable low-rank matrices, namely a down-projection and an up-projection matrix, each of which is followed by a feature transformation vector. Based on a novel \textit{freezing score}, we then incrementally freeze these projection matrices during fine-tuning to reduce the computation and alleviate over-fitting. Our experimental results demonstrate that we can achieve state-of-the-art performance with an average improvement of up to $0.85\%$ as evaluated on the GLUE benchmark while yielding up to $9.5\times$ fewer average trainable parameters. While compared in terms of runtime, AFLoRA can yield up to $1.86\times$ improvement as opposed to similar PEFT alternatives. Besides the practical utility of our approach, we provide insights on the trainability requirements of LoRA paths at different modules and the freezing schedule for the different projection matrices. The code will be released.

\end{abstract}

\section{Introduction}
Pre-trained language models such as BERT \citep{devlin2018bert}, GPT-3 \citep{gpt3}, and LLaMA2 \citep{touvron2023llama} have demonstrated commendable performance on various natural language processing (NLP) tasks \cite{kang2024gear}. However, their zero-shot performance on many downstream tasks often falls short of expectations. One possible solution is full fine-tuning (FFT) of the model on the downstream dataset. However, the large model parameter size makes this process prohibitively costly. \\

\begin{figure}
\centerline{\includegraphics[scale=0.28]{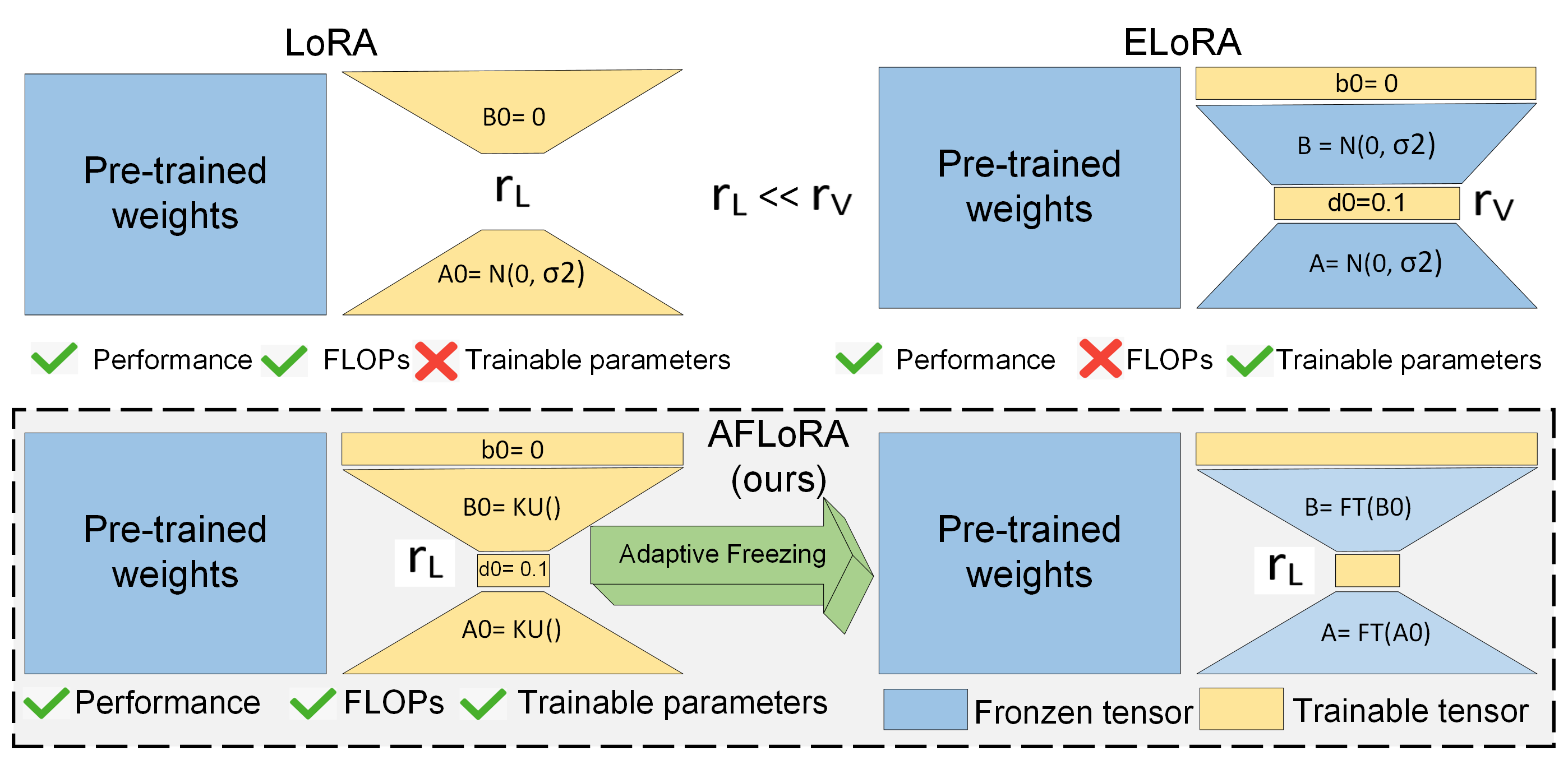}}
\caption{
 Schematic comparison of LoRA \citep{hu2021lora}, ELoRA \citep{kopiczko2024elora}, and AFLoRA and their associated advantages and disadvantages in terms of various metrics. $r_L$ and $r_V$, represent the rank of the low-rank path used in LoRA and ELoRA methods, respectively. FT and KU refer to fine-tuned weights and the Kaiming uniform initialization function, respectively.}
 \vspace{-6mm}
\label{fig:aflora_framework}
\end{figure}

To address this challenge, various \textit{parameter-efficient fine-tuning} (PEFT) methods including low rank adaptation (LoRA) \citep{hu2021lora}, adapter tuning \citep{he2021effectiveness}, and prompt tuning \citep{lester2021power} are proposed. These methods add parameters to the trained model for fine-tuning, 
bypassing the need to adjust the weights of the pre-trained model. 
In particular, LoRA \citep{hu2021lora} and its variants \citep{zhang2023adaptive} add a 
trainable low-rank path consisting of down-projection and up-projection matrices to the model, inspired by \cite{aghajanyan2020intrinsic} which showed that such low-rank paths can effectively approximate the trained weight tensors. ELoRA \cite{kopiczko2024elora} extends LoRA by adding trainable feature transformation vectors to the output of each project matrix. They showed that SoTA accuracy can be achieved with the projection matrices frozen after random initialization while keeping the two feature transformation vectors trainable. This approach significantly reduces the number of trainable parameters. However, compared to LoRA, ELoRA incurs higher computation costs due to the higher rank needed for the frozen projection 
matrices. Fig. \ref{fig:aflora_framework} illustrates LoRA and ELoRA, contrasting them to our proposed method AFLoRA.\\

\textbf{Our contributions.} To reduce the trainable parameter count and computation costs of fine-tuning, we present \textit{Adaptive Freezing of Low Rank Adaptation} (AFLoRA). More specifically, we first investigate the rank needed for the frozen LoRA path in ELoRA and observe that reducing the rank of the frozen projection matrices (PM) causes a drop in fine-tuning performance. 

Based on this insight, we present AFLoRA, which starts with a low-rank trainable path that includes projection matrices and feature transformation vectors and trains the path for some epochs. We then gradually freeze the projection matrices based on a novel \textit{freezing score} that acts as a proxy for the trainability requirement of a LoRA tensor. In this way, we not only help alleviate the over-fitting issue but also, improve the computation efficiency. To evaluate the benefit of AFLoRA, we perform extensive evaluations on multiple NLP benchmark datasets and compare accuracy, FLOPs, and training time with 
several existing alternatives. Specifically, compared to ELoRA we yield $1.86\times$ and $2.96\times$ improvement in runtime and FLOPs, respectively, while remaining comparable as LoRA on these two metrics. Compared to LoRA we require $9.5\times$ fewer average trainable parameters to yield similar or improved performance.  

\section{Related Works}
PEFT \citep{hu2021lora, kundu2024sensi, sridhar2023instatune, yin2024pruning} refers to a collection of methodologies that focus on allowing a small number of parameters to fine-tune to yield good performance on a downstream task. For example, prefix-tuning \citep{li2021prefix} adds trainable prefix tokens to a model's input or hidden layers while adapter-tuning \citep{houlsby2019adapter} inserts small neural network layers, known as adapters, within each layer of a pre-trained model. LoRA \citep{hu2021lora}, on the other hand, adds low-rank tensors in parallel to the frozen pre-trained weights. AdaLoRA \citep{zhang2023adaptive} allows the rank of the LoRA path to be chosen in an adaptive way.  Other variants like SoRA \cite{ding2023sparse} and LoSparse \cite{li2023losparse} have investigated the impact of sparsity in and alongside the low-rank path, respectively.  Recently, efficient low-rank adaptation (ELoRA) \citep{kopiczko2024elora} has proposed to keep the LoRA path frozen, while introducing two trainable feature transformation vectors. Thus, this work only studies an extreme scenario of keeping the LoRA path frozen, and, to the best of our knowledge, no work has investigated 
the trainability requirement of the projection matrices.\\

\section{Motivational Case Study}
\begin{figure}
\centerline{\includegraphics[scale=0.08]{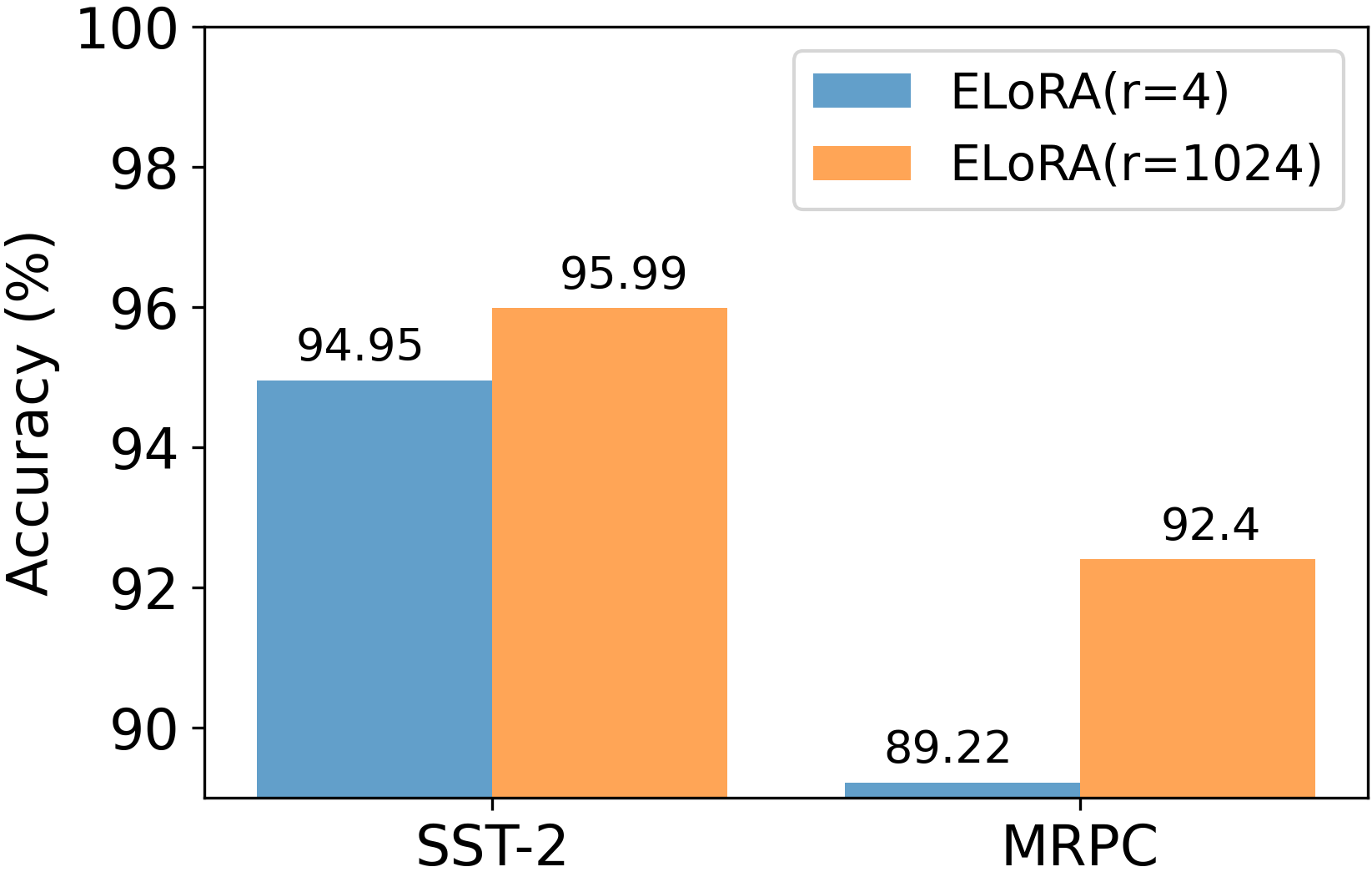}}
\caption{
 Performance of ELoRA with two different ranks of the frozen projection matrices.}
 \vspace{-10mm}
\label{fig:elora_motive}
\end{figure}

To understand the high-rank requirement for the frozen projection matrices in ELoRA, we conduct two sets of fine-tuning on SST-2 and MRPC, with ELoRA having rank ($r$) of 1024 and 4, respectively. As we can see in Fig. \ref{fig:elora_motive}, the model with $r=4$, yields poorer performance, highlighting the need for high rank for the frozen tensors. This high rank causes ELoRA to potentially be FLOPs inefficient.

\section{AFLoRA: Methodology}
\begin{table*}[!t]
\tiny\addtolength{\tabcolsep}{-1.2pt}
\label{table:comparison_on_glue}
\centering
\caption{Comparison of different LoRA variants with DeBERTaV3 on the GLUE benchmark.}
\vspace{-3mm}
\begin{tabular}{l|c|ccccccccc}
\hline
Method & \#Params. $\downarrow$ & CoLA $\uparrow$ & SST-2 $\uparrow$ & MRPC $\uparrow$ & QNLI $\uparrow$ & STS-B $\uparrow$ & RTE $\uparrow$ & MNLI $\uparrow$ & QQP $\uparrow$ & Avg. $\uparrow$ \\
\hline
FFT & 184M & 69.21 & 95.64 & 89.22 & 93.78 & 91.59 & 82.49 & 89.98/89.95 & 92.05/89.31 & 87.82 \\
LoRA (r = 8) & 1.33M & 69.73 & 95.57 & 89.71 & 93.76 & \textbf{91.86} & 85.32 & \textbf{90.47/90.46} & 91.95/89.26 & 88.38 \\
AdaLoRA & 1.27M & 70.86 & 95.95 & 90.22 & 94.28 & 91.39 & 87.36 & 90.27/90.30 & \textbf{92.13}/88.41 & 88.83 \\
SoRA (r = 4) & 0.47M & 71.05 & 95.57 & 90.20 & 93.92 & 91.76 & 86.04 & 90.38/90.43 & 92.06/\textbf{89.44} & 88.71\\
ELoRA* & 0.16M & 70.74 & 95.18 & 90.93 & 93.58 & 91.08 & 87.36 & 90.11/90.22 & 90.69/87.63 & 88.53 \\
\rowcolor{Gray}
AFLoRA (r = 4) & \textbf{0.14M}** & \textbf{72.01} & \textbf{96.22} & \textbf{91.91} & \textbf{94.42} & 91.84 & \textbf{88.09} & 89.88/90.17 & 90.81/87.77 & \textbf{89.23} \\
\hline
\end{tabular}
\begin{tablenotes}
    \tiny{
        \item * The original paper has results with the RoBERTa,  we generated the results with our implementation on DeBERTaV3 with rank of 1024.   
        \item ** As the number of trainable parameters is changed during training, we computed this by averaging over the whole training epochs over all datasets. 
    }
\end{tablenotes}
\vspace{-4mm}

\end{table*}

\textbf{Module Structure.}
Inspired by the framework proposed by \citet{kopiczko2024elora}, we design the LoRA module to encompass four components, namely, the down-projection linear layer ($lora_A$), the up-projection linear layer ($lora_B$), and two feature transform vectors ($s_d$, and $s_b$) placed before and after $lora_B$. However, unlike \cite{kopiczko2024elora},  \textbf{we keep both the projection matrices ($lora_A$ and $lora_B$) and vectors trainable at the beginning and keep the rank very low}.
The module processes a given input $X$ through these components to produce an output $Y$. The complete operation for a layer $l$ can be described as follows:
\begin{equation}
    {Y} = {W^l_0}{X} + \Lambda^l_b B^l \Lambda^l_d A^l{X}
    \label{eq:elora}
\end{equation}

Here, $A^l$ and $ B^l$ are the trainable LoRA tensors of $lora^l_A$ and $lora^l_B$, respectively. $\Lambda_d$ and $\Lambda_b$ are the vectors of $s_d$, and $s_b$, respectively. $W^l_0$ represents the frozen pre-trained weights.
We use Kaiming Uniform initialization for $A^l$ and $ B^l$, and follow \cite{kopiczko2024elora} to initialize the vectors.\\

\textbf{Adaptive Freezing.}
\label{sec:freezing score}
In pruning literature \citep{han2015learning, molchanov2019importance, zhang2022platon, yin2024pruning, kundu2021dnr, kundu2022bmpq}, sensitivity is gauged to reflect weight variability, necessitating consideration of both the weights' magnitudes and their gradients. Small weight values suggest minimal impact, while minor gradient values indicate stability. Taking inspiration from this idea, here we introduce the concept of a "freezing score". However, unlike pruning where both magnitude and gradient play a critical role in identifying insignificant weight, we leverage only gradient as a proxy to compute the freezing score. This is because, we assume large magnitude weights with negligible change has the same priority to be frozen as that for small magnitude weights. This score quantifies the degree to which weights vary throughout the training process. Consequently, when the expected changes to the weights become negligible, we may consider them to be frozen, thereby saving computational resources and energy.

The following equation describes the freezing score evaluation steps for a low-rank tensor $A^l$.

{\small
\begin{equation} \label{eq:ipt:1}
    I_{A^l} = \left| \nabla \mathcal{L}(\boldsymbol{\theta)} \right|, \overline{I}_{A^l}^{(t)} = \beta_1 \overline{I}_{A^l}^{(t-1)} + (1 - \beta_1) I_{A^l}^{(t)}
\end{equation}
\vspace{-2pt}
\begin{equation}\label{eq:ipt:2}
    U_{A^l}^{(t)} = \left| I_{A^l}^{(t)} - \overline{I}_{A^l}^{(t)} \right|,
    \overline{U}_{A^l}^{(t)} = \beta_2 \overline{U}_{A^l}^{(t-1)} + (1 - \beta_2) U_{A^l}^{(t)}
\end{equation}
\vspace{-2pt}
\begin{equation}\label{eq:ipt:3}
    s_{A^l}^{(t)} = mean( \overline{I}_{A^l}^{(t)} \circ \overline{U}_{A^l}^{(t)})
\end{equation}
}
Here, for each projection tensor at iteration $t$, we compute a smoothed gradient ($\overline{I}_{A^l}^{(t)}$) and uncertainly tensor ($\overline{U}_{A^l}^{(t)}$), as shown in Eq. 2 and 3, respectively. We then evaluate the freezing score  $s_{A^l}^{(t)}$, as the mean of the tensor generated via Hadamard product ($\circ$) between $\overline{I}_{A^l}^{(t)}$ and $\overline{U}_{A^l}^{(t)}$.
% in accordance with the methodology outlined in Eq. \ref{eq:ipt}. 

To apply thresholding on the LoRA freezing scores, we use the cubic schedule as \citep{zhang2022platon}. In specific, we keep the projection matrices trainable for the initial $t_i$ training steps, and then progressively freeze them by calculating the freezing fraction $r(t)$ as shown in the Eq. \ref{eq:cubic}.
Finally, all the projection matrices freeze beyond $T - t_f$ steps. Note, at step $t$, for a computed freezing fraction $k$, we freeze the lowest $k\%$ projection matrices.  

{\small
\vspace{-2pt}
\begin{equation}\label{eq:cubic}
    r(t) = \left\{
  \begin{array}{ll}
  0 & \quad 0 \leq t < t_i \\
  1 -\left(1 - \frac{t-t_i}{T-t_i-t_f}\right)^3 & \quad t_i \leq t < T - t_f \\
  1 & \quad \text{otherwise}
  \end{array}
\right.
\end{equation}
\vspace{-2pt}
}

where $t$ refers to current \#step,  $T$ is the total number of fine-tuning steps.
We set $t_i$ to the steps corresponding to one epoch and set $t_f$ to 70\% of the total training steps.

\section{Experiments}
\textbf{Models \& Datasets.} We use the PEFT framework of \cite{peft} and evaluate the fine-tuning performance of DeBERTaV3-base \citep{he2020deberta} to fine-tune on our framework on the General Language Understanding Evaluation (GLUE) benchmark \citep{wang2018glue}. The details of the hyperparameter settings for each dataset are listed in Appendix \ref{apdx:hyper}. \\

\textbf{Performance Comparison.}
We benchmark the performance with AFLoRA and present comparison with LoRA and its variants. For ELoRA, we reproduce the results at our end while the results for other methods are sourced from \cite{ding2023sparse}. As shown in Table 1, AFLoRA can achieve SoTA performance on the majority of datasets and on average while requiring similar and $9.5\times$ fewer average trainable parameters as compared to ELoRA and LoRA, respectively. 
\\

\begin{figure}
\centerline{\includegraphics[scale=0.13]{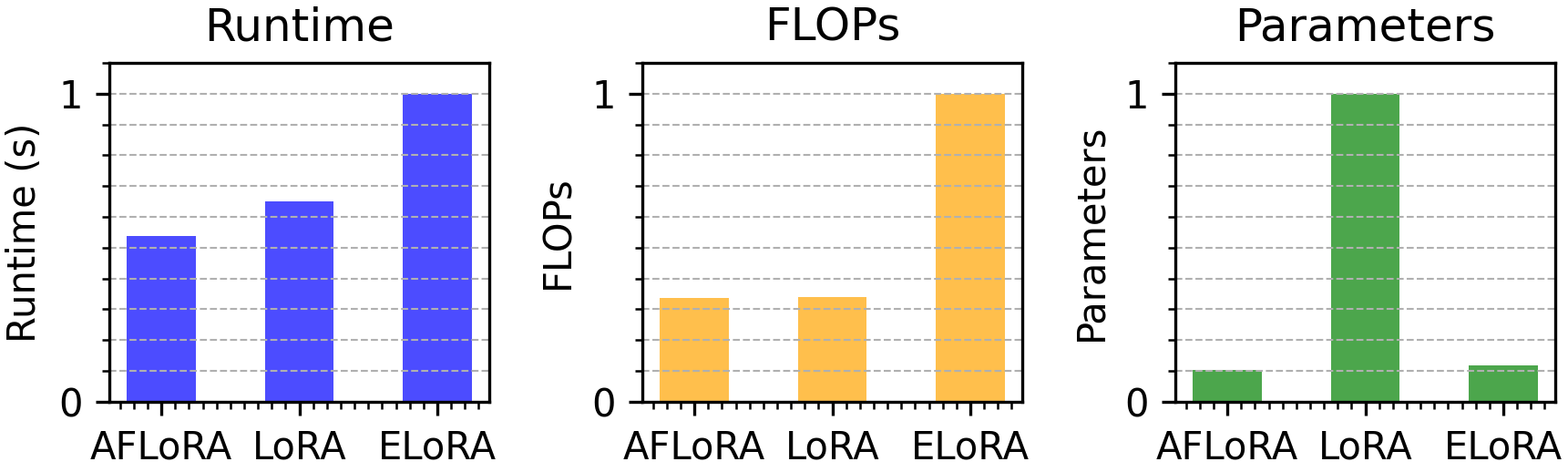}}
\caption{
A comparison of various system performances between LoRA, ELoRA, and AFLoRA.}
\vspace{-7mm}
\label{fig:system}
\end{figure}

\textbf{Runtime \& FLOPs  Comparison.}
Fig. \ref{fig:system} shows the comparison of the normalized average training runtime, normalized FLOPs, and normalized trainable parameters. For AFLoRA, we average the training time, FLOPs, and trainable parameters over six GLUE datasets (except the MNLI and QQP datasets). Note, for LoRA and ELoRA, the trainable parameters and FLOPs remain fixed for all the datasets. We compute their average runtime the same way as ours. Compared to ELoRA we can yield up to $1.86\times$ and $2.96\times$ runtime and FLOPs improvement while remaining comparable with LoRA in these two metrics. Compared to LoRA we yield $9.5\times$ parameter reduction, while remaining comparable with ELoRA. These results clearly demonstrate AFLoRA
as PEFT method that can yield similar parameter efficiency as ELoRA while costing no training overhead in FLOPs or time. \\

\textbf{Results with Large Language Models (LLMs).} We now demonstrate the AFLoRA fine-tuning performance with two popular LLM variants, namely, LLaMA-7B \cite{touvron2023llama} and BART-Large \cite{lewis2019bart} on GSM8k complex reasoning and CNN/Daily mail summarizing task, respectively. As demonstrated in Table 2, on GSM8k, AFLoRA yields improved accuracy of $1.09\%$ while requiring $3.15\times$ fewer trainable parameters as compared to that with LoRA. On CNN/DailyMail Summarizing task (Table 3), AFLoRA requires $1.69\times$ fewer trainable parameters to reach similar or improved rouge score values.

\begin{table}[t]
\tiny\addtolength{\tabcolsep}{-3.0pt}
%\vspace{-1mm}
\begin{center}
{
%\begin{threeparttable}
\label{table:llm_ft_result}
\caption{Results on Auto regressive complex reasoning task using LLM. }
\vspace{-4mm}
\begin{tabular}{l|c|c|c|c}
\hline
Method & Model & Low-rank val. & \# Params. & GSM8k Acc (\%) \\
\hline
LoRA  & LLaMA-7B & 32 & 56.1M & 37.50 \\
\rowcolor{Gray} AFLoRA (Ours) &  LLaMA-7B & 32 & \textbf{17.8M} & \textbf{38.59} \\ 
\hline
\end{tabular}
%\end{threeparttable}
}
\vspace{-2mm}
\end{center}
\end{table}

\begin{table}[t]
\tiny\addtolength{\tabcolsep}{-3.0pt}
%\vspace{-1mm}
\label{table:llm_ft_result_cnn}
\begin{center}
{
\begin{threeparttable}
\caption{Results on Summarizing task using LLM. We use rouge 1 (R1) and rouge 2 (R2) score to measure the summarization quality.}
\vspace{-4mm}
\begin{tabular}{l|c|c|c|c}
\hline
Method & Model & Low-rank val. & \# Params. & CNN/DailyMail (R1/R2) \\
\hline
LoRA  & BART-Large & 16 & 8.65M & 43.96/21.06\\
\rowcolor{Gray} AFLoRA (Ours) &  BART-Large & 16 & \textbf{5.10M} & \textbf{44.31/21.32} \\ 
\hline
\end{tabular}
\end{threeparttable}
}
\vspace{-2mm}
\end{center}
\end{table}

\begin{table}[t]
\tiny\addtolength{\tabcolsep}{-3.0pt}
%\vspace{-1mm}
\label{table:adaptive_freezing}
\begin{center}
{
\begin{threeparttable}
\caption{Ablation study on the trainability impact of the projection matrices (PM) of the AFLoRA module. We keep the vectors trainable throughout for all. }
\vspace{-4mm}
\begin{tabular}{l|c|ccccccc}
\hline
PM & \#Params. & CoLA & SST-2 & MRPC & QNLI & STS-B & RTE & Avg. \\
\hline
Trainable  & 0.45M & 70.15 & 95.99 & \textbf{92.4} & 94.16 & 89.90 & \textbf{88.45} & 88.51 \\
Frozen   & 0.08M & 70.36 & 94.95 & 89.22 & 93.61 & 91.17 & 85.92 & 87.54 \\
\rowcolor{Gray} AFLoRA (Ours) & 0.14M & \textbf{72.01} & \textbf{96.22} & 91.91 & \textbf{94.42} & \textbf{91.84} & 88.09 & \textbf{89.23} \\ 
\hline
\end{tabular}
\end{threeparttable}
}
\vspace{-3mm}
\end{center}
\end{table}

\section{Ablations and Discussions}
We conducted our ablation studies on six GLUE benchmark datasets, omitting QQP and MNLI, the two most computationally demanding datasets.\\

\textbf{Do we really need adaptive freezing?}
We conducted experiments with all the LoRA PMs frozen (same as ELoRA), all the LoRA PMs trainable, and with our adaptive training of LoRA PMs. We use, $r=4$ for the LoRA path, for all. As we can see in Table 4, keeping the projection matrices trainable yields better average performance compared to keeping them frozen throughout. However, AFLoRA with adaptive freezing yields even better performance than keeping them trainable throughout, potentially highlighting its ability to regularize the fine-tuning against overfitting.\\

% \vspace{-3mm}
\textbf{Do we need to keep the PMs trainable for all layer types?}
There are two major layer types, FFN and the attention layers. We place the PMs in both along with the feature transformation vectors. We then study the necessity of keeping the PMs trainable in these two layer types. Note, here, we keep the vectors trainable for all throughout. As shown in Table 5, keeping the PMs trainable (and then adaptive freezing) in the FFN yields better performance compared to the alternatives. Note we keep the PMs in the attention layers frozen to random values. Interestingly, allowing all PMs to initially train and then adaptively freeze yields poorer performance than allowing them only in MLP. This may hint at the FFN weights to play a more important role in fine-tuning performance. \\

\begin{figure}[!t]
\centerline{\includegraphics[scale=0.40]{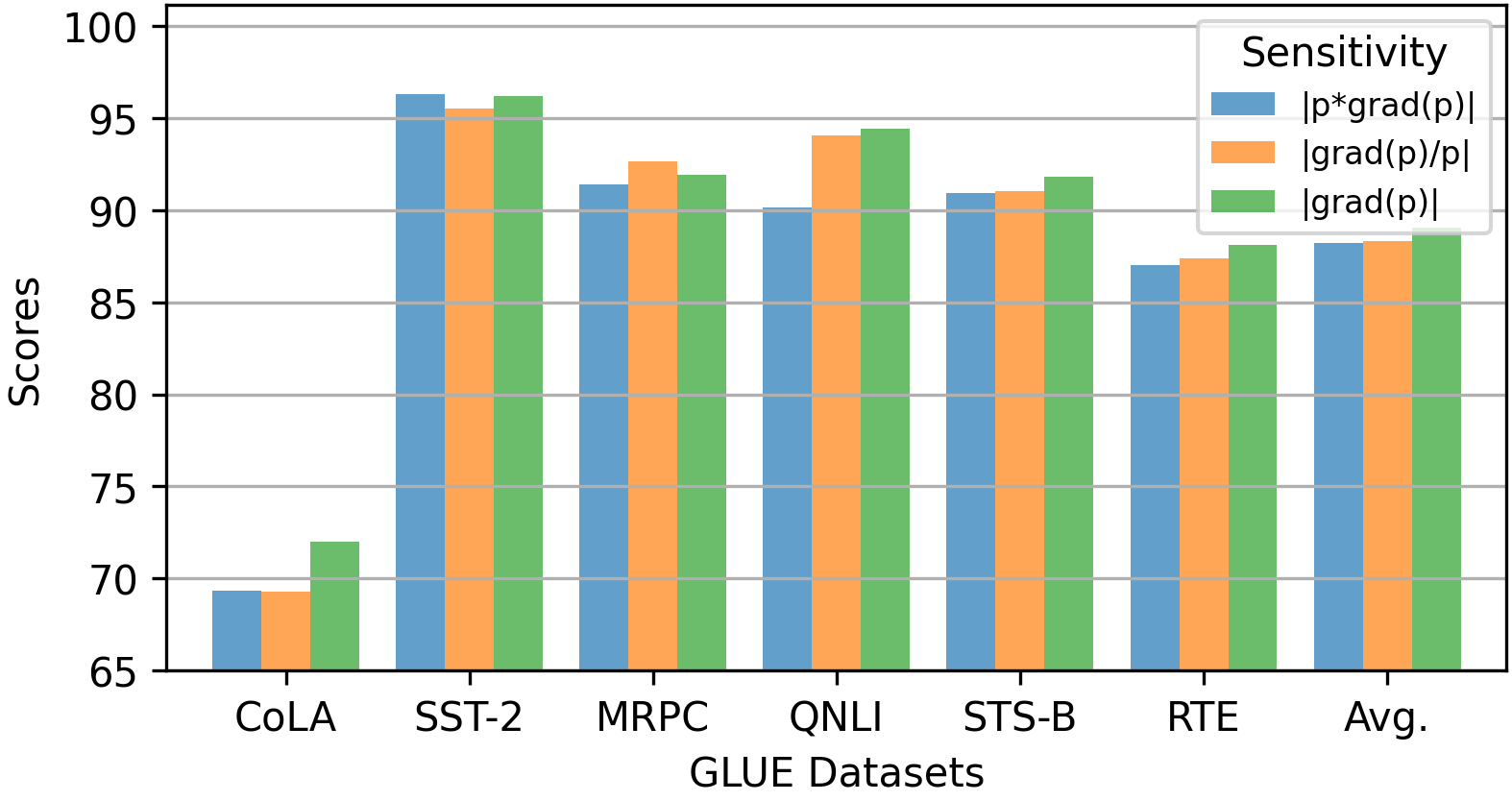}}
\vspace{-4mm}
\caption{
A comparison of performance outcomes utilizing three distinct freezing score methodologies.}
\vspace{-2mm}
\label{fig:ipts}
\end{figure}

\begin{table}[!t]
\tiny\addtolength{\tabcolsep}{-1.5pt}
\caption{Ablation study on making the PMs for different layer-types trainable.}
\vspace{-3mm}
\label{tab:ab_diff_part}
\begin{center}
{
\begin{threeparttable}
\begin{tabular}{l|l|ccccccc}
\hline
FFN & Attn & CoLA & SST-2 & MRPC & QNLI & STS-B & RTE & Avg. \\
\hline
\cmark & \cmark & 70.33 & 95.76 & 90.93 & 94.36 & 91.44 & 87.37 & 88.48 \\
 & & 0.15M  & 0.19M & 0.18M & 0.19M & 0.16M & 0.17M & 0.17M\\
\hline
\xmark & \cmark & 71.118 & 95.986 & 89.951 & 94.12 & 91.39 & 86.28 & 88.14  \\
 &  & 0.11M  & 0.13M & 0.12M & 0.13M & 0.12M & 0.12M & 0.12M\\
 \hline
\rowcolor{Gray} \cmark & \xmark  & \textbf{72.01} & \textbf{96.22} & \textbf{91.91} & \textbf{94.42} & \textbf{91.84} & \textbf{88.09} & \textbf{89.02} \\ 
& & 0.13M & 0.18M & 0.13M & 0.13M & 0.13M & 0.13M &  0.14M \\
\hline
\end{tabular}
\end{threeparttable}
}
\vspace{-5mm}
\end{center}
\end{table}

\textbf{Ablation with sensitivity choices.}
Fig. \ref{fig:ipts} presents ablation with three sensitivity scores based on three different sensitivity choices, namely, $|grad(p)|$ (adopted in AFLoRA), $|p*grad(p)|$, and $|grad(p)/p|$. On average, the freezing score adopted in AFLoRA, consistently yields better accuracy over the other two. \\

\begin{figure}[!t]
\centerline{\includegraphics[scale=0.46]{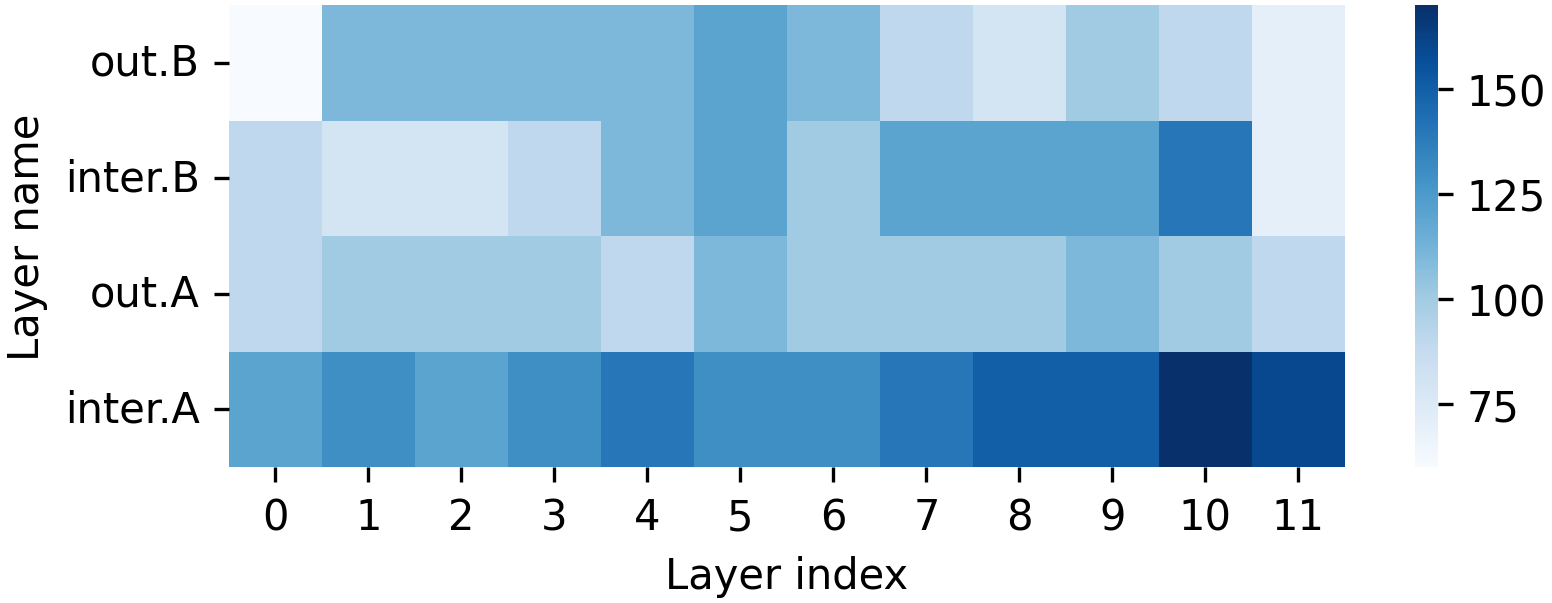}}
\vspace{-3mm}
\caption{
 Visualization of freezing iterations for each layer. `out' and `inter' refer to the second and the first MLP layer of the FFN, respectively. `A' and `B' represent the down-projection and up-projection matrix, respectively. The darker the color, the more iterations the matrix has to go through before freezing.}
 \vspace{-6mm}
\label{fig:heatmap}
\end{figure}

\textbf{Discussion on Freezing Trend.}
We use the RTE dataset as a case study, to understand the freezing trend of the PMs across different layers. Specifically, we illustrate the specific number of iterations required before freezing each component in Fig. \ref{fig:heatmap}. Interestingly, as can be seen from the figure, analysis reveals that the down-projection matrix parallel to the intermediate linear layer requires longer training duration prior to being frozen, as compared to the other PMs. This may potentially hint at the low approximation ability of the intermediate layer as compared to the second MLP in the FFN.

\section{Conclusions}
In this paper, we presented AFLoRA, an adaptive freezing of LoRA adapters that allow near-optimal trainability of the LoRA projection matrices and freezes them driven by a "freezing score" after certain fine-tuning steps. Compared to LoRA, AFLoRA can reduce the trainable parameters by up to $9.5\times$ while yielding $0.85\%$ average improved performance as evaluated on the GLUE benchmark.

\section{Limitation}
In the ablation study with various freezing score metrics, we discovered that alternative scoring methods outperform ours on certain datasets, suggesting possible room for research in refining the freezing scores. This can further improve performance with AFLoRA. Additionally, the integration of AFLoRA in the adaptive rank evaluation framework can potentially open a new direction of PEFT that we consider as future research.

\bibliography{anthology,custom}
\bibliographystyle{acl_natbib}

\newpage
\appendix
\section{Appendix}
\label{sec:appendix}

\subsection{Dataset}
The details of train/test/dev splits and the evaluation metric of the GLUE \citep{wang2018glue} dataset are reported in Table \ref{tab:glue_data_stats}. We use the Huggingface Transformers library \cite{wolf-etal-2020-transformers} to source all the datasets.
\begin{table}[h]
\tiny\addtolength{\tabcolsep}{-2.5pt}
\centering
\caption{Statistics of the GLUE benchmark datasets."Mcc", "Acc", "F1" and "Pear" represent Matthews correlation coefficient, accuracy, the
F1 score and the Pearson correlation coefficient respectively. And "Acc" for MNLI dataset contains the accuracy for the matched and mismatched subset of the datasets. }
\vspace{-3mm}
\begin{tabular}{lcccc}
\hline
Dataset & \#Train & \#Valid & \#Test & Metric \\
\hline
CoLA & 8.5k & 1,043 & 1,063 & Mcc \\
SST-2 & 67k & 872 & 1.8k & Acc \\
MRPC & 3.7k & 408 & 1.7k & Acc \\
QQP & 364k & 40.4k & 391k & Acc/F1 \\
STS-B & 5.7k & 1.5k & 1.4k & Pear \\
MNLI & 393k & 9.8k/9.8k & 9.8k/9.8k & Acc \\
QNLI & 105k & 5.5k & 5.5k & Acc \\
RTE & 2.5k & 277 & 3k & Acc \\
\hline
\end{tabular}

\label{tab:glue_data_stats}
\end{table}
\vspace{-2mm}
\subsection{Hyperparameter configuration}
\label{apdx:hyper}
Table \ref{tab: hyper} shows the main hyper-parameter set up in this paper. Besides them, we use the same optimizer, warmup Ratio, and LR schedule as \citet{hu2021lora}. We use NVIDIA RTX A6000 (maximum GPU memory=49140MB) to measure the training runtime.
For all experiments, we run 5 times using
different random seeds and report the average
results.
\begin{table}[!h]
\tiny\addtolength{\tabcolsep}{-3.0pt}
\caption{Hyperparameter setup for all eight datasets in GLUE benchmark}
\vspace{-3mm}
\label{tab: hyper}
\begin{center}
{
\begin{threeparttable}
\begin{tabular}{l|cccccccc}
\hline
Hyperparameter & CoLA & SST-2 & MRPC & QNLI & STS-B & RTE & MNLI & QQP  \\
\hline
\# epochs & 20 & 10 & 20 & 10 & 20 & 20 & 10 & 10\\
Batch size & \multicolumn{8}{c}{64} \\
Max Seq. Len. & \multicolumn{8}{c}{256} \\
Clf. Lr.* & 4E-2 & 4E-3 & 8E-2 & 4E-3 & 2E-2 & 4E-2 & 4E-3 & 4E-3 \\
Learning rate & 1E-2 & 4E-3 & 1E-2 & 1E-3 & 2E-3 & 1E-3 & 1E-3 & 4E-3 \\
\hline
$t_i (epoch)$ & \multicolumn{8}{c}{1} \\
$t_f (epoch)$ & 14 & 7 & 14 & 7 & 14 & 14 & 7 & 7 \\
$\beta_1$ & \multicolumn{8}{c}{0.85}\\
$\beta_2$ & \multicolumn{8}{c}{0.95}\\
\hline
\end{tabular}
\begin{tablenotes}
    \footnotesize{
        \item * "Clf. Lr.*" means the learning rate for the classification head.   
    }
\end{tablenotes}
\end{threeparttable}
}
\vspace{-5mm}
\end{center}
\end{table}

\subsection{Ablation study on if freezing the two projection matrices in the same layer simultaneously}
We study the value of freezing both projection matrices in the same layer simultaneously. The results, depicted in Table \ref{tab:ab:shareAB}, demonstrate that freezing the projection matrices separately yields consistently superior performance compared to freezing them simultaneously.
% \cite{li2023loftq}, \cite{hu2023llm}, \cite{touvron2023llama}, \cite{touvron2023llama2}

\begin{table}[!t]
\small\addtolength{\tabcolsep}{-2.5pt}
    \centering
    \caption{Ablation study on whether freezing the two projection matrices in the same layer simultaneously or independently.}
    \vspace{-3mm}
    \begin{tabular}{c|cc}
    \hline
         & Simultaneously & Independently \\
         \hline
        CoLA & 67.90 & 72.01 \\
        SST-2 & 95.87 & 96.22 \\
        MRPC & 91.67 & 91.91 \\
        STS-B & 91.64 & 91.84 \\
        QNLI & 94.20 & 94.42 \\
        RTE & 87.00 & 88.09 \\
        Avg. & 88.05 & 89.02 \\
        \#Params & 0.146M & 0.138M \\
        \hline
    \end{tabular}
    \label{tab:ab:shareAB}
\end{table}

\end{document}